\documentclass{article}
\usepackage{ijcai26}

\usepackage{times}
\usepackage{soul}
\usepackage{url}
\usepackage[hidelinks]{hyperref}
\usepackage[utf8]{inputenc}
\usepackage[small]{caption}
\usepackage{graphicx}
\usepackage{tikz}
\usetikzlibrary{arrows.meta,positioning}
\usepackage{amsmath}
\usepackage{amsthm}
\usepackage{booktabs}
\usepackage{algorithm}
\usepackage[switch]{lineno}
\usepackage{tcolorbox}

\usepackage[T1]{fontenc}    
\usepackage{amsfonts}       
\usepackage{nicefrac}       
\usepackage{microtype}      
\usepackage[table,xcdraw]{xcolor}

\usepackage{algorithmicx}
\usepackage{algpseudocode}
\usepackage{xspace}
\usepackage{mathrsfs}
\usepackage{multirow}
\usepackage{enumitem}
\usepackage{subcaption}
\usepackage{pifont}
\usepackage{threeparttable}
\usepackage{color}
\usepackage{balance} 
\usepackage{amssymb}
\usepackage{bm}
\usepackage{makecell}

\usepackage{framed}

\title{When LLMs Analyze Scars: From Images to Clinically-Meaningful Features}

\author{Ruman Wang\textsuperscript{\normalfont 1}, Hangting Ye\textsuperscript{\normalfont 2}\\
Liaoning University of Traditional Chinese Medicine\textsuperscript{\normalfont 1} \\School of Artificial Intelligence, Jilin University\textsuperscript{\normalfont 2}\\ 
\texttt{ruman.wang@outlook.com}, \texttt{yeht22@mails.jlu.edu.cn}
}

\begin{document}

\maketitle

\begin{abstract}
Medical image classification faces a fundamental dilemma: while deep learning models achieve remarkable performance at scale, real-world clinical scenarios often suffer from severe data scarcity due to annotation costs, privacy constraints, and disease rarity. This challenge is particularly pronounced in pathological scar classification, where differentiating keloids from hypertrophic scars requires subtle expert knowledge and labeled images are extremely limited.
We propose a novel paradigm that repositions large language models (LLMs) as \emph{knowledge-driven feature engineers} rather than end-to-end classifiers. We call this framework \textbf{ScaFE} (\textbf{Sc}ar \textbf{F}eature \textbf{E}ngineering). Our key insight is that LLMs encode rich medical knowledge that can be externalized as executable feature extraction code, enabling the transformation of high-dimensional images into low-dimensional, clinically interpretable representations. Specifically, we prompt an LLM with established scar assessment criteria to generate deterministic Python code that extracts features aligned with clinical scoring systems such as the Vancouver Scar Scale.
Our approach offers three key advantages: (1) \textbf{data efficiency}, achieving robust performance with limited training samples by decoupling knowledge acquisition from statistical learning; (2) \textbf{privacy preservation}, as raw images are processed locally without exposure to external LLMs; and (3) \textbf{interpretability}, through explicit features grounded in clinical reasoning. Extensive experiments on scar classification demonstrate that our method consistently outperforms end-to-end deep learning baselines or using LLMs as black-box classifiers under limited data conditions, establishing a promising direction for integrating LLMs into data-efficient and clinically transparent medical AI systems.
\end{abstract}

\section{Introduction}

Pathological scars, including keloids and hypertrophic scars, represent a significant clinical challenge affecting millions of patients worldwide~\cite{bayat2003skin,sidgwick2015comprehensive}. Although these two conditions share similar visual appearances, their underlying pathophysiology, prognosis, and treatment strategies differ substantially: keloids extend beyond the original wound boundaries and exhibit high recurrence rates, while hypertrophic scars remain confined to the wound site and often regress spontaneously~\cite{berman2017keloids}. Accurate differentiation is therefore essential for appropriate treatment planning; however, diagnosis requires substantial expert experience and remains subject to considerable inter-observer variability.

From a machine learning perspective, automated scar classification exemplifies a broader challenge in medical AI: the tension between model complexity and data availability. State-of-the-art deep neural networks demand large-scale labeled datasets to learn robust representations, yet such data are scarce in specialized medical domains. Acquiring high-quality scar annotations requires board-certified dermatologists or plastic surgeons, making large-scale labeling prohibitively expensive. Privacy regulations further restrict data sharing across institutions, resulting in small, heterogeneous datasets that violate the assumptions underlying data-hungry models. Consequently, end-to-end neural networks trained directly on raw images often overfit to spurious correlations and fail to generalize in clinical deployment~\cite{litjens2017survey}.

Recent advances in large language models (LLMs) offer a promising alternative pathway. Modern LLMs encode vast amounts of medical knowledge distilled from clinical literature, textbooks, and practice guidelines~\cite{singhal2023large,nori2023capabilities}. Unlike conventional neural networks that learn features purely from data, LLMs can articulate expert-level clinical reasoning and operationalize established diagnostic criteria. However, directly applying multimodal LLMs to medical images introduces practical concerns: sensitive patient data must be transmitted to external servers, model outputs lack reproducibility due to inherent stochasticity, and the black-box nature of such systems undermines clinical trust.

These observations motivate our central question: \emph{How can we leverage the rich clinical knowledge embedded in LLMs for medical image understanding without directly exposing raw images or relying on large-scale end-to-end training?}

Inspired by the diagnostic reasoning process of human clinicians, we observe that expert physicians rarely make diagnoses based solely on holistic visual impressions. Instead, they systematically translate visual patterns into structured, clinically meaningful concepts, such as pigmentation abnormality, surface texture, vascular prominence, and boundary characteristics, before synthesizing these observations into a diagnostic conclusion. This cognitive process naturally suggests a decomposition: first extract interpretable features that capture clinical semantics, then perform classification in this structured feature space.

Building on this insight, we propose to use LLMs as \emph{knowledge-based feature engineers} that bridge the gap between raw medical images and structured clinical reasoning. Rather than training models to learn features from data, we prompt a medical LLM to generate executable Python code that extracts clinically meaningful features according to established scar assessment criteria such as the Vancouver Scar Scale (VSS) and the Patient and Observer Scar Assessment Scale (POSAS)~\cite{busche2018burn}. The generated code defines a deterministic mapping from images to interpretable feature vectors, which are then classified using lightweight machine learning models.

This design resolves the fundamental tension between model capacity and data availability. By externalizing domain knowledge into the feature construction stage, our approach enables robust learning even with limited training samples. Since feature extraction is performed locally through generated code, patient images never leave the clinical environment. Moreover, the extracted features are directly interpretable and aligned with clinical vocabulary, facilitating transparent decision-making and expert validation.

Our contributions can be summarized as follows:
\begin{itemize}[leftmargin=*,itemsep=2pt,topsep=2pt]
    \item We propose \textbf{ScaFE} (\textbf{Sc}ar \textbf{F}eature \textbf{E}ngineering), a framework that leverages LLMs as knowledge-driven feature engineers, transforming medical images into clinically meaningful structured representations via generated executable code.
    \item We demonstrate that ScaFE with lightweight classifiers achieves superior performance in low-data scar classification settings, outperforming end-to-end deep learning baselines.
    \item We provide an interpretable and privacy-preserving alternative to conventional medical image analysis, with features explicitly grounded in established clinical assessment scales.
\end{itemize}

\section{Related Work}

\subsection{Automated Scar Assessment}
Computational approaches to scar assessment have evolved from handcrafted feature engineering to end-to-end deep learning. Early methods relied on features derived from clinical scoring systems such as the Vancouver Scar Scale~\cite{baryza1995vancouver} and the Patient and Observer Scar Assessment Scale~\cite{draaijers2004posas}, extracting color, texture, and morphological descriptors that mirror human assessment criteria. While interpretable, these approaches required extensive manual design and exhibited limited robustness across diverse imaging conditions.

Recent work has applied convolutional neural networks to scar classification and severity prediction~\cite{zhu2016deep}, achieving promising results under controlled experimental settings. However, these data-driven approaches assume access to large, well-annotated datasets---an assumption that rarely holds in clinical practice where labeled scar images are scarce and expensive to obtain. Furthermore, the end-to-end nature of deep learning models renders their decision-making opaque, hindering clinical adoption where interpretability is paramount.

\subsection{Large Language Models in Medical Imaging}
The emergence of large language models has opened new possibilities for medical image analysis. Multimodal LLMs such as GPT-4V~\cite{achiam2023gpt} and Med-PaLM~\cite{singhal2023large} have demonstrated impressive capabilities in interpreting medical images and generating clinically relevant explanations. Several studies have explored using vision-language models for dermatological diagnosis~\cite{liu2023derm,zhou2024pre}, radiology report generation~\cite{chen2024medcap}, and general medical visual question answering~\cite{li2023llava}.

Despite their strong reasoning capabilities, directly applying multimodal LLMs to clinical diagnosis presents significant challenges. First, transmitting sensitive patient images to external API servers raises privacy concerns under regulations such as HIPAA~\cite{gostin2009beyond}. Second, the stochastic nature of LLM outputs compromises reproducibility, a critical requirement in medical applications. Third, these models often function as black boxes, providing explanations that may not align with established clinical frameworks. Our work addresses these limitations by using LLMs as code generators rather than direct classifiers, preserving privacy while ensuring deterministic and interpretable feature extraction.

\subsection{Learning from Structured Representations}
Learning from structured, low-dimensional representations has long been recognized as effective in data-limited settings~\cite{bengio2013representation}. Classical approaches combine domain knowledge with statistical learning through feature engineering, rule-based systems, or hybrid architectures~\cite{domingos2012few}. Recent work has revisited this paradigm by integrating symbolic representations with neural networks, demonstrating improved sample efficiency and interpretability~\cite{garcez2019neural,ye2024ptarl}.

In medical AI, structured representations derived from clinical knowledge have shown particular promise. Concept bottleneck models~\cite{koh2020concept} learn intermediate clinical concepts before final prediction, enabling human intervention and interpretation. Knowledge-guided neural networks incorporate medical ontologies and clinical guidelines into the learning process~\cite{xie2019knowledge}. Our approach extends this line of research by using LLMs as automated generators of clinically grounded feature extractors, eliminating the need for manual feature design while ensuring alignment with established medical criteria.

\subsection{Positioning of Our Work}
Our method occupies a unique position in the landscape of medical image analysis. Unlike end-to-end deep learning, we do not require large-scale training data, as domain knowledge is externalized through LLM-generated code. Unlike direct LLM-based classification, we preserve privacy by processing images locally and ensure reproducibility through deterministic feature extraction. Unlike manual feature engineering, our approach automatically translates clinical criteria into executable code, reducing human effort while maintaining interpretability. Table~\ref{tab:positioning} summarizes these distinctions.

\begin{table}[t]
\centering
\caption{Comparison of our approach with existing paradigms.}
\label{tab:positioning}
\small
\begin{tabular}{lccccc}
\toprule
Method & Data Eff. & Privacy & Interpret. & Auto. \\
\midrule
End-to-End DL & \ding{55} & \ding{51} & \ding{55} & \ding{51} \\
Multimodal LLM & \ding{51} & \ding{55} & \ding{55} & \ding{51} \\
Manual Features & \ding{51} & \ding{51} & \ding{51} & \ding{55} \\
\textbf{Ours} & \ding{51} & \ding{51} & \ding{51} & \ding{51} \\
\bottomrule
\end{tabular}
\end{table}

\section{Problem Formulation}

Let $\mathcal{I} \subset \mathbb{R}^{H \times W \times 3}$ denote the space of RGB clinical scar images, and let $\mathcal{Y} = \{1, \ldots, C\}$ represent the set of $C$ scar categories (\emph{e.g.}, keloid and hypertrophic scar). Given a labeled training dataset
\begin{equation}
\mathcal{D}_{\text{train}} = \{(I_i, y_i)\}_{i=1}^{N}, \quad I_i \in \mathcal{I}, \; y_i \in \mathcal{Y},
\end{equation}
where $N$ is typically small due to annotation costs and privacy constraints, our goal is to learn a classifier $f: \mathcal{I} \rightarrow \mathcal{Y}$ that accurately predicts scar categories for unseen test images.

\paragraph{Challenges of End-to-End Learning.}
The conventional approach trains a high-capacity neural network $f_{\theta}: \mathcal{I} \rightarrow \mathcal{Y}$ by minimizing empirical risk over $\mathcal{D}_{\text{train}}$. However, when $N$ is small relative to the model capacity, this leads to overfitting: the learned features capture dataset-specific artifacts rather than clinically meaningful patterns. Moreover, the black-box nature of such models hinders interpretability and clinical trust.

\paragraph{Proposed Decomposition.}
To address these challenges, we decompose the classification task into two stages:
\begin{enumerate}[leftmargin=*,itemsep=2pt]
    \item \textbf{Feature Extraction:} A mapping $\phi: \mathcal{I} \rightarrow \mathbb{R}^K$ that transforms raw images into a $K$-dimensional feature space capturing clinically meaningful attributes.
    \item \textbf{Lightweight Classification:} A simple classifier $h: \mathbb{R}^K \rightarrow \mathcal{Y}$ trained on the structured feature representations.
\end{enumerate}
The composite classifier is then $f = h \circ \phi$, with prediction given by $\hat{y} = h(\phi(I))$.

The key insight is that designing $\phi$ to encode \emph{clinical knowledge} substantially reduces the effective hypothesis space, enabling robust learning from limited data. Rather than learning features from scratch, we leverage prior medical expertise to construct representations that align with established diagnostic criteria.

\paragraph{Design Requirements for $\phi$.}
An effective feature extractor $\phi$ for our setting should satisfy:
\begin{itemize}[leftmargin=*,itemsep=2pt]
    \item \textbf{Clinical Meaningfulness:} Features should correspond to attributes used by clinicians (\emph{e.g.}, erythema, texture, boundary characteristics).
    \item \textbf{Determinism:} Given the same input image, $\phi$ should produce identical outputs, ensuring reproducibility.
    \item \textbf{Privacy Preservation:} Feature extraction should be performed locally without transmitting raw images externally.
\end{itemize}

In the following section, we describe how we leverage large language models to automatically construct $\phi$ by generating executable feature extraction code grounded in established scar assessment criteria.

\section{Methodology}

\subsection{Overview}

Our framework explicitly decouples domain knowledge acquisition from statistical learning. While end-to-end neural networks entangle feature learning and classification within a single optimization process, we separate these stages by introducing an LLM-guided feature construction module that injects expert knowledge \emph{prior} to learning.

\begin{figure*}[t]
\centering
\includegraphics[width=0.9\textwidth, height=7.2cm]{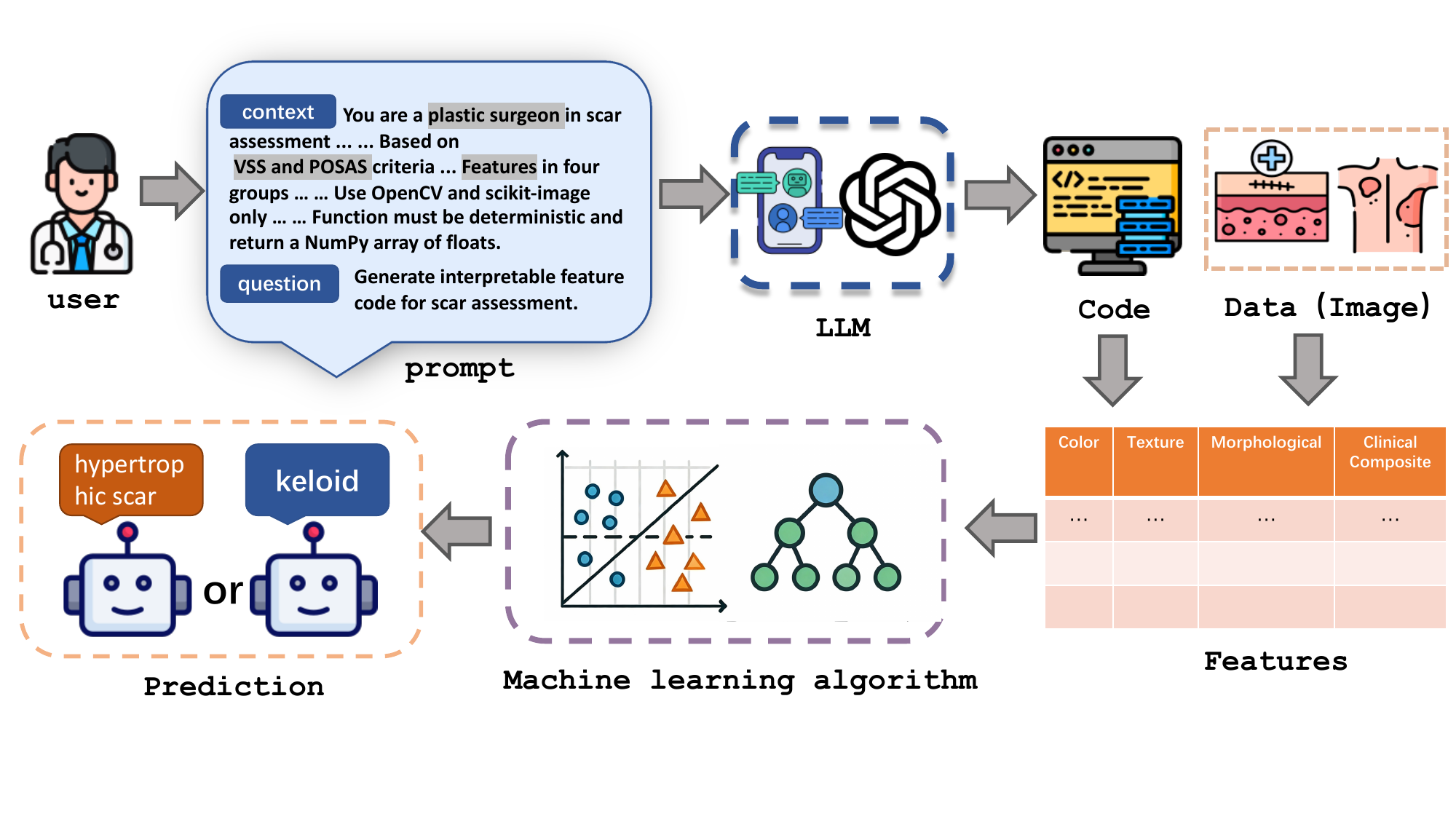}
\caption{Overview of the proposed LLM-guided feature engineering framework. An LLM generates feature extraction code based on clinical criteria, which transforms scar images into structured representations for classification.}
\label{fig:framework}
\end{figure*}

As illustrated in Figure~\ref{fig:framework}, the framework consists of three components: (i) \textbf{LLM-based code generation}, which externalizes medical knowledge into executable Python code; (ii) \textbf{structured feature extraction}, which deterministically maps images to interpretable clinical features; and (iii) \textbf{lightweight classification}, which performs statistical learning in the low-dimensional feature space.

\subsection{LLM-Guided Feature Code Generation}

\paragraph{Prompt Design.}

We design a structured prompt $\mathcal{P}$ to elicit clinically grounded feature extraction code from an LLM. The prompt comprises three components:
\begin{equation}
\mathcal{P} = \mathcal{P}_{\text{role}} \oplus \mathcal{P}_{\text{medical}} \oplus \mathcal{P}_{\text{code}},
\end{equation}
where $\oplus$ denotes concatenation and:
\begin{itemize}[leftmargin=*,itemsep=2pt]
    \item $\mathcal{P}_{\text{role}}$ establishes the expert persona (\emph{e.g.}, ``You are a board-certified plastic surgeon specializing in scar assessment...'').
    \item $\mathcal{P}_{\text{medical}}$ is accomplished by the assistance of clinicians. It encodes domain knowledge by referencing established clinical scales including the Vancouver Scar Scale (VSS) and Patient and Observer Scar Assessment Scale (POSAS). It also specifies the expected feature categories, which is detailed in Section~\ref{sec: Clinical support.}.
    \item $\mathcal{P}_{\text{code}}$ specifies the output format, including the Python function signature, required libraries (OpenCV, scikit-image).
\end{itemize}

\paragraph{Prompt Template.}
We use a structured prompt to instruct the LLM to generate deterministic feature extraction code. An example template is in Template~\ref{lst:prompt}:
\begin{tcolorbox}[
    colframe=black,
    colback=gray!10,
    coltitle=white,
    colbacktitle=black,
    title=Prompt Template,
    fonttitle=\footnotesize\bfseries,
    fontupper=\scriptsize,
    arc=2mm,
    boxrule=0.5pt,
    top=1mm, bottom=1mm, left=2mm, right=2mm,
    label={lst:prompt}
]
\textbf{\textcolor{blue}{$\mathcal{P}_{\text{role}}$:}} You are a board-certified plastic surgeon specializing in scar assessment. 
\textbf{\textcolor{blue}{$\mathcal{P}_{\text{medical}}$:}} Use the Vancouver Scar Scale (VSS) and POSAS criteria to design interpretable image features. Include features in four groups: color (LAB statistics), texture (LBP, gradients, entropy), morphology (solidity, circularity, elongation), and clinical composite scores.
\textbf{\textcolor{blue}{$\mathcal{P}_{\text{code}}$:}} Output a Python function \texttt{extract\_features(image)} that returns a fixed-length vector. Use OpenCV and scikit-image only. The function must be deterministic and return a NumPy array of floats.
\end{tcolorbox}

\paragraph{Code Generation.}

Given an LLM with parameters $\Theta$, we obtain the feature extraction code:
\begin{equation}
\label{eq:code}
\mathcal{G} = \text{LLM}_{\Theta}(\mathcal{P}) \in \mathcal{C}_{\text{Python}},
\end{equation}
where $\mathcal{C}_{\text{Python}}$ denotes the space of valid Python programs. The generated code $\mathcal{G}$ defines a deterministic function that maps each image $\mathcal{I}$ to feature vectors $\mathbf{F} \in \mathbb{R}^K$:
\begin{equation}
\label{eq:code apply}
\mathcal{G}: \mathcal{I} \rightarrow \mathbb{R}^K, \quad I \mapsto \mathbf{F} = [F_1, F_2, \ldots, F_K]^\top,
\end{equation}
where each element of $\mathbf{F}$ is a clinical-meaningful feature. Crucially, the code generation process via LLM (Eq.~\ref{eq:code}) is data-agnostic, and it preserves the patient privacy information without exposing raw scar images to the LLM, where LLM reasons over its own encoded clinical knowledge. In addition, this process yields a reusable code without requiring any LLM fine-tuning, which can then be used to process all the patient samples locally (Eq.~\ref{eq:code apply}).

\paragraph{Code Validation and Refinement.}

To ensure robustness, we validate the generated code through: (i) syntax checking, (ii) execution on sample images, and (iii) verification of output dimensionality. If validation fails, we provide error feedback to the LLM for iterative refinement until valid code is obtained. In practice, we find that well-designed prompts typically yield valid code within one or two iterations.

\subsection{Clinical Assurance and Case Study}
\label{sec: Clinical support.}
\paragraph{Clinical Assurance.}
We co-designed the prompt with a board-certified plastic surgeon specializing in scar assessment. The expert reviewed the clinical criteria (VSS/POSAS), verified the terminology, and confirmed that the requested features reflect routine clinical practice. This ensures the generated code encodes clinically meaningful measurements rather than arbitrary image statistics.

\subsection{Clinical Assurance and Case Study}
\label{sec: Clinical support.}
\paragraph{Clinical Assurance.}
We co-designed the prompt with a board-certified plastic surgeon specializing in scar assessment. The expert confirmed that the four feature groups correspond to clinical criteria used in VSS/POSAS: vascularity and pigmentation (color), surface roughness and pliability (texture), boundary spread and elevation (morphology), and holistic severity (composites). This ensures the extractor operationalizes clinically meaningful cues rather than arbitrary image statistics.

\paragraph{Case Study: From Image $\mathcal{I}$ to Features.}
To illustrate how the feature extractor behaves on a representative scar image, we now present a case study for the generated code $\mathcal{G}$.
Given an RGB scar image $\mathcal{I} \in \mathbb{R}^{H \times W \times 3}$, the code $\mathcal{G}$ first isolates a scar region via a mask:  $\Omega \subset \{1,\dots,H\}\times\{1,\dots,W\}$ (by simple thresholding and morphological cleanup on color contrast) and then computes features on $\Omega$.
The extracted features can be summarized into four clinically interpretable groups:
\begin{equation}
\mathbf{F} = [\phi^{\text{color}}, \phi^{\text{texture}}, \phi^{\text{morph}}, \phi^{\text{clin}}]^\top \in \mathbb{R}^K,
\end{equation}
with each group defined below.

\noindent\textbf{\textit{Color Features $\phi^{\text{color}}$.}}
Code $\mathcal{G}$ transforms $\mathcal{I}$ to CIELAB, where each pixel has three channels: $L^*$ (lightness), $a^*$ (red–green axis), and $b^*$ (yellow–blue axis). Each channel is a scalar field in $\mathbb{R}^{H \times W}$. Over $\Omega$, code $\mathcal{G}$ computes:
\begin{equation}
\phi^{\text{color}} = \left[ \frac{\mu_a - \mu_{a,\text{ref}}}{\sigma_{a,\text{ref}}}, \; \frac{\mu_L - \mu_{L,\text{ref}}}{\sigma_{L,\text{ref}}}, \; \text{Var}(a^*), \; \text{Skew}(b^*) \right].
\end{equation}
Here $\mu_a$ and $\mu_L$ are the mean $a^*$ and $L^*$ values within $\Omega$, while $(\mu_{a,\text{ref}},\sigma_{a,\text{ref}})$ and $(\mu_{L,\text{ref}},\sigma_{L,\text{ref}})$ are computed from a nearby normal-skin region. The normalized terms measure how much the scar’s redness/brightness deviates from healthy skin. $\text{Var}(a^*)$ is the variance of $a^*$ over $\Omega$, and $\text{Skew}(b^*)$ is the third standardized moment (skewness) of $b^*$, capturing asymmetry in pigmentation distribution.

\noindent\textbf{\textit{Texture Features $\phi^{\text{texture}}$.}}
Let $I_g \in \mathbb{R}^{H \times W}$ be the grayscale image. Code $\mathcal{G}$ computes:
\begin{equation}
\phi^{\text{texture}} = \left[ \text{LBP}_{8,1}^{\text{uniform}}, \; \frac{1}{|\Omega|}\sum_{p \in \Omega}|\nabla I_g(p)|, \; \text{Entropy}(I_g) \right].
\end{equation}
$\text{LBP}_{8,1}^{\text{uniform}}$ is a uniform local binary pattern histogram statistic (radius 1, 8 neighbors). $\nabla I_g(p)$ is the gradient at pixel $p$, and the mean gradient magnitude quantifies surface irregularity. $\text{Entropy}(I_g)$ is the Shannon entropy of grayscale intensities in $\Omega$, measuring texture complexity. Clinically, these reflect scar roughness and pliability.

\noindent\textbf{\textit{Morphological Features $\phi^{\text{morph}}$.}}
From $\Omega$, code $\mathcal{G}$ computes area $A$ (pixel count), perimeter $P$ (boundary length), convex hull area $A_{\text{convex}}$, and ellipse axes $(\lambda_{\text{major}}, \lambda_{\text{minor}})$ fitted to $\Omega$:
\begin{equation}
\phi^{\text{morph}} = \left[ \frac{A}{A_{\text{convex}}}, \; \frac{P^2}{4\pi A}, \; \frac{\lambda_{\text{major}}}{\lambda_{\text{minor}}} \right].
\end{equation}
$A/A_{\text{convex}}$ (solidity) decreases when scars extend irregularly; $P^2/(4\pi A)$ (circularity) captures boundary irregularity; and axis ratio quantifies elongation. These correspond to scar spread beyond the wound boundary.

\noindent\textbf{\textit{Clinical Composite Features $\phi^{\text{clin}}$.}}
Code $\mathcal{G}$ aggregates clinically relevant cues into composite scores:
\begin{equation}
\phi^{\text{clin}} = \left[ \sigma(\beta^\top [\phi^{\text{color}}, \phi^{\text{texture}}]), \; \widehat{\text{VSS}} \right],
\end{equation}
where $\sigma$ is the sigmoid function, $\beta \in \mathbb{R}^{d}$ is a weight vector that controls the contribution of each color/texture feature to the composite severity score, and $\widehat{\text{VSS}}$ is a VSS-like severity proxy computed from the extracted cues.

\paragraph{Clinical Agreement.}
The clinician reviewed the feature definitions and confirmed that each group corresponds to routine assessment criteria in VSS/POSAS, and that the normalization against nearby healthy skin makes the measurements interpretable and clinically meaningful.

\subsection{Classification Pipeline}

Algorithm~\ref{alg:pipeline} summarizes the complete training and inference procedure.

\begin{algorithm}[t]
\caption{LLM-Guided Feature Engineering Pipeline}
\label{alg:pipeline}
\begin{algorithmic}[1]
\Require Training data $\mathcal{D}_{\text{train}} = \{(I_i, y_i)\}_{i=1}^N$, test image $I_{\text{test}}$, LLM with parameters $\Theta$
\Ensure Predicted label $\hat{y}_{\text{test}}$
\Statex \textbf{// Stage 1: Feature Code Generation (One-time)}
\State Construct prompt $\mathcal{P} = \mathcal{P}_{\text{role}} \oplus \mathcal{P}_{\text{medical}} \oplus \mathcal{P}_{\text{code}}$
\State Generate code: $\mathcal{G} \leftarrow \text{LLM}_{\Theta}(\mathcal{P})$
\State Validate and refine $\mathcal{G}$ if necessary
\Statex \textbf{// Stage 2: Feature Extraction}
\For{$i = 1$ to $N$}
    \State $\mathbf{F}_i \leftarrow \mathcal{G}(I_i)$ \Comment{Apply generated code}
\EndFor
\State Construct feature dataset: $\mathcal{D}^*_{\text{train}} = \{(\mathbf{F}_i, y_i)\}_{i=1}^N$
\Statex \textbf{// Stage 3: Classifier Training}
\State Train lightweight classifier $h$ on $\mathcal{D}^*_{\text{train}}$
\Statex \textbf{// Stage 4: Inference}
\State Extract test features: $\mathbf{F}_{\text{test}} \leftarrow \mathcal{G}(I_{\text{test}})$
\State Predict: $\hat{y}_{\text{test}} \leftarrow h(\mathbf{F}_{\text{test}})$
\State \Return $\hat{y}_{\text{test}}$
\end{algorithmic}
\end{algorithm}

\paragraph{Training Phase.}
Applying the generated code $\mathcal{G}$ to training images yields a structured dataset $\mathcal{D}^*_{\text{train}} = \{(\mathbf{F}_i, y_i)\}_{i=1}^N$. We train a lightweight classifier $h: \mathbb{R}^K \rightarrow \mathcal{Y}$, such as Support Vector Machine~\cite{suthaharan2016support} and Random Forest~\cite{breiman2001random}, on this low-dimensional representation. Hyperparameters are selected via cross-validation on the training set.

\paragraph{Inference Phase.}
For a test image $I_{\text{test}}$, we first extract features $\mathbf{F}_{\text{test}} = \mathcal{G}(I_{\text{test}})$ using the same generated code, then obtain the prediction $\hat{y}_{\text{test}} = h(\mathbf{F}_{\text{test}})$.

\subsection{Discussion: Why This Works}

Our approach succeeds by addressing the core challenge of medical image classification: the mismatch between model complexity and data availability. We highlight three key mechanisms:

\paragraph{Knowledge Externalization.}
Rather than learning clinical concepts from limited data, we leverage the LLM's pre-existing medical knowledge to construct features that already encode relevant diagnostic criteria. This dramatically reduces the effective hypothesis space that the downstream classifier must search.

\paragraph{Dimensionality Reduction.}
The feature extractor maps high-dimensional images ($H \times W \times 3 \approx 10^5$--$10^6$ dimensions) to compact representations ($K \approx 10$--$20$ dimensions). This massive compression, guided by clinical relevance, enables effective learning even with small $N$.

\paragraph{Deterministic Processing.}
Unlike direct LLM-based classification, our generated code produces identical outputs for identical inputs, ensuring reproducibility. The code can be inspected, validated, and modified by domain experts, enhancing trust and enabling human-in-the-loop refinement.

\section{Experiments}

We conduct comprehensive experiments to evaluate our proposed method, addressing the following research questions:
\begin{itemize}[leftmargin=*,itemsep=1pt,topsep=2pt]
    \item \textbf{RQ1:} Does our method outperform baselines in low-data scar classification?
    \item \textbf{RQ2:} How does performance scale with varying training set sizes?
    \item \textbf{RQ3:} What is the contribution of each feature category?
    \item \textbf{RQ4:} Is the framework stable across multiple LLM generation runs and different LLMs?
    \item \textbf{RQ5:} Are the generated features interpretable?
\end{itemize}

\subsection{Experimental Setup}



\paragraph{Dataset.}
We focus on binary classification (keloid (KD) vs. hypertrophic scar (HS)) only. We collected patient images from our partner hospitals with informed consent, and all data were approved for research use. Data collection and expert labeling took 10 months, so the cost was very high and only 40 high quality images are available (20 KD, 20 HS). 

\paragraph{Evaluation Protocol.}
Given the limited dataset size, we adopt stratified 5-fold cross-validation and report mean and standard deviation across folds. All results are averaged over three runs with different random seeds.
We report Accuracy, Sensitivity (Recall), Specificity, and F1-score, which are standard metrics in medical image analysis. 

\paragraph{Implementation Details.}
The LLM-guided feature extractor is generated once using GPT-4 with temperature 0 to maximize determinism. The generated code uses OpenCV and scikit-image for image processing. For downstream classification, we evaluate SVM (RBF kernel), Random Forest and Decision Tree classifier with inner cross-validation on the training set. Deep baselines are trained with PyTorch on two NVIDIA RTX 4090 GPUs using Adam (lr=$1\times10^{-4}$), random horizontal flip and color jitter. ResNet18 and EfficientNet-B0 start from ImageNet weights; ViT-Base is trained from scratch.

\subsection{Baselines}

We compare against representative methods spanning different paradigms:

\begin{itemize}[leftmargin=*,itemsep=2pt]
    \item \textbf{CNN-ResNet18}: ResNet-18~\cite{he2016deep} pretrained on ImageNet, fine-tuned on scar images.
    \item \textbf{CNN-EfficientNet}: EfficientNet-B0~\cite{tan2019efficientnet} pretrained on ImageNet, fine-tuned on scar images.
    \item \textbf{ViT-Base}: Vision Transformer~\cite{dosovitskiy2020image} trained from scratch on scar images.
    \item \textbf{Handcrafted+RF}: Classical features (color histograms, GLCM texture, shape descriptors) with Random Forest classifier.
    \item \textbf{MLLM-Direct}: GPT-4V prompted to directly classify scar images~\cite{shiraishi2024potential} without code generation.
\end{itemize}

MLLM-Direct uses GPT-4V for zero-shot classification on anonymized images with numeric labels; we report single-run results when available to avoid label leakage.
For deep learning baselines, we apply standard data augmentation (random crop, flip, color jitter) and early stopping to mitigate overfitting.

\subsection{Main Results (RQ1)}

Table~\ref{tab:main_results} presents classification performance between \textbf{ScaFE} (\textbf{Sc}ar \textbf{F}eature \textbf{E}ngineering) and baselines on the binary scar classification task. Our method consistently outperforms handcrafted features and remains competitive with transfer-learned deep models under the low-data regime.

\begin{table}[t]
\centering
\caption{Performance comparison on binary scar classification (keloid vs. hypertrophic). \textbf{ScaFE} (SVM) indicates that the downstream classifier $h$ used in our proposed \textbf{ScaFE} is SVM. Best results in \textbf{bold}, second-best \underline{underlined}. Results are mean $\pm$ std over 5 folds.}
\label{tab:main_results}
\scriptsize
\setlength{\tabcolsep}{4pt}
\resizebox{\linewidth}{!}{%
\begin{tabular}{lcccc}
\toprule
Method & Acc (\%) & Sens. & Spec. & F1 \\
\midrule
CNN-ResNet18 & 0.61 $\pm$ 0.13 & 0.62 $\pm$ 0.13 & 0.62 $\pm$ 0.13 & 0.55 $\pm$ 0.17 \\
CNN-EfficientNet & 0.58 $\pm$ 0.14 & 0.58 $\pm$ 0.14 & 0.58 $\pm$ 0.14 & 0.55 $\pm$ 0.15 \\
ViT-Base & 0.51 $\pm$ 0.04 & 0.50 $\pm$ 0.00 & 0.50 $\pm$ 0.00 & 0.34 $\pm$ 0.02 \\
MLLM-Direct & 0.60 $\pm$ 0.24 & 0.60 $\pm$ 0.00 & 0.60 $\pm$ 0.00 & 0.56 $\pm$ 0.00 \\
Handcrafted+RF & 0.66 $\pm$ 0.20 & 0.65 $\pm$ 0.20 & 0.65 $\pm$ 0.20 & 0.64 $\pm$ 0.21 \\ \midrule
\textbf{ScaFE} (SVM) & 0.68 $\pm$ 0.17 & 0.68 $\pm$ 0.16 & 0.68 $\pm$ 0.16 & 0.67 $\pm$ 0.17 \\
\textbf{ScaFE} (DT) & \underline{0.69 $\pm$ 0.17} & \underline{0.69 $\pm$ 0.16} & \underline{0.69 $\pm$ 0.16} & \underline{0.68 $\pm$ 0.18} \\
\textbf{ScaFE} (RF) & \textbf{0.73 $\pm$ 0.19} & \textbf{0.72 $\pm$ 0.18} & \textbf{0.72 $\pm$ 0.18} & \textbf{0.72 $\pm$ 0.19} \\
\bottomrule
\end{tabular}
}
\end{table}

Several observations emerge. First, our LLM-guided features substantially outperform handcrafted baselines, confirming that clinically grounded feature generation captures more relevant cues than manual design. Second, while ResNet18 benefits from ImageNet pretraining, its performance comes at the cost of heavier models and reduced interpretability. Third, all three of our variants outperform the baselines. Since RF classifier achieves the best performance among our variants, we adopt RF for subsequent analyses (RQ2--RQ4).

\subsection{Data Efficiency Analysis (RQ2)}

\begin{figure}[t]
\centering
\includegraphics[width=0.8\linewidth]{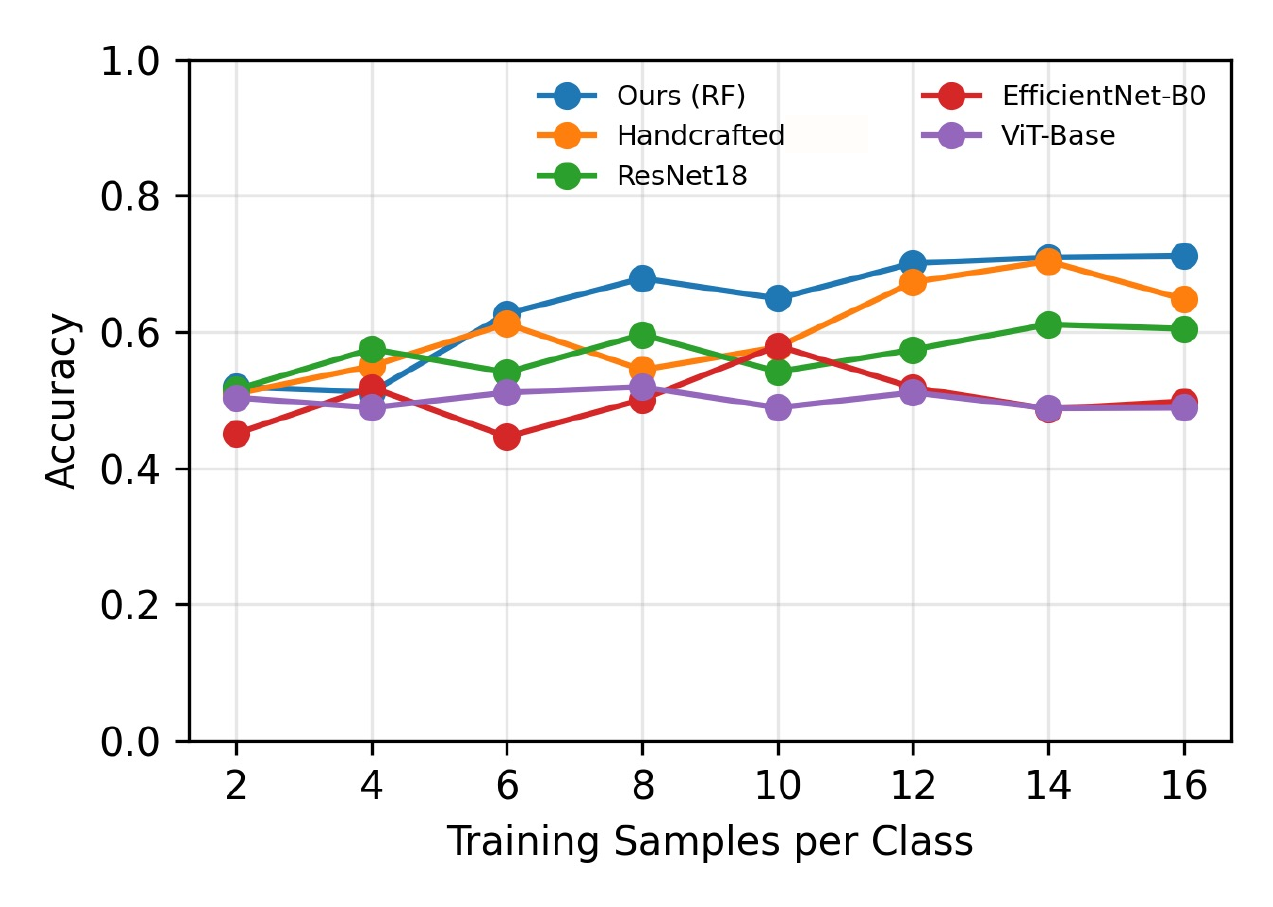}
\caption{Classification accuracy under different training data sizes. Our method maintains robust performance even with severely limited data.}
\label{fig:data_efficiency}
\end{figure}

To investigate robustness under varying data availability, we subsample the training set to fixed per-class counts of {2, 4, 6, 8, 10, 12, 14, 16} images. Figure~\ref{fig:data_efficiency} compares ScaFE (ours) with handcrafted features and deep baselines under the same splits.
As shown in Figure~\ref{fig:data_efficiency}, ScaFE achieves the best overall performance, and is robust across all shots. When the training set size drops significantly to 2 samples per class, the performance does not drop significantly, highlighting robust few-shot behavior.

\subsection{Ablation Studies (RQ3)}

\paragraph{Feature Construction Strategy.}
We isolate the contribution of LLM-guided feature engineering by comparing against alternative feature construction strategies.
\begin{table}[t]
\centering
\caption{Ablation study on feature construction strategies.}
\label{tab:ablation_feature}
\scriptsize
\setlength{\tabcolsep}{4pt}
\resizebox{\linewidth}{!}{%
\setlength{\tabcolsep}{5.0mm}{
\begin{tabular}{lcc}
\toprule
Feature Type & Acc (\%) & F1 \\
\midrule
Random Features & 0.69 $\pm$ 0.16 & 0.69 $\pm$ 0.18 \\
Handcrafted (No LLM) & 0.66 $\pm$ 0.20 & 0.64 $\pm$ 0.21 \\
LLM w/o Medical Prompt & 0.68 $\pm$ 0.16 & 0.66 $\pm$ 0.17 \\
LLM-Generated (Ours) & 0.73 $\pm$ 0.19 & 0.72 $\pm$ 0.19 \\
\bottomrule
\end{tabular}
}}
\end{table}
Results in Table~\ref{tab:ablation_feature} show that LLM-guided features outperform handcrafted baselines, and removing medical context from the prompt reduces performance. 

\paragraph{Feature Group Contribution.}
We analyze the contribution of each feature category by selectively removing one group at a time.
\begin{table}[t]
\centering
\caption{Ablation on feature groups. All features combined achieves the best performance.}
\label{tab:ablation_group}
\scriptsize
\setlength{\tabcolsep}{4pt}
\resizebox{\linewidth}{!}{%
\setlength{\tabcolsep}{5.0mm}{
\begin{tabular}{lcc}
\toprule
Configuration & Acc (\%) & F1 \\
\midrule
All Features & 0.73 $\pm$ 0.19 & 0.72 $\pm$ 0.19 \\
w/o Color Features & 0.70 $\pm$ 0.16 & 0.68 $\pm$ 0.17 \\
w/o Texture Features & 0.71 $\pm$ 0.16 & 0.71 $\pm$ 0.16 \\
w/o Morph Features & 0.64 $\pm$ 0.17 & 0.61 $\pm$ 0.20 \\
w/o Clin Features & 0.71 $\pm$ 0.14 & 0.70 $\pm$ 0.14 \\
\bottomrule
\end{tabular}
}}
\end{table}
Table~\ref{tab:ablation_group} shows that removing morphological features causes the largest drop, while clinical composite has a moderate effect, indicating morphology carries the most discriminative signal for keloid vs. hypertrophic scars. This is aligned with clinical experience.

\subsection{Stability Analysis (RQ4)}

Although LLMs exhibit inherent stochasticity, our framework mitigates this by: (i) using low temperature during generation, and (ii) producing deterministic executable code rather than free-form text outputs. We assess stability by running the feature generation process five times with identical prompts across GPT-4 and Gemini-2.5.
\begin{table}[t]
\centering
\caption{Stability analysis (Accuracy) across different LLMs (GPT-4 and Gemini-2.5) with multiple generation runs.}
\label{tab:stability}
\scriptsize
\setlength{\tabcolsep}{4pt}
\resizebox{\linewidth}{!}{%
\setlength{\tabcolsep}{7.0mm}{
\begin{tabular}{lccc}
\toprule
Run & GPT-4 & Gemini-2.5 \\
\midrule
Run 1 & 0.73 $\pm$ 0.19 & 0.72 $\pm$ 0.16 \\
Run 2 & 0.73 $\pm$ 0.21 & 0.74 $\pm$ 0.19 \\
Run 3 & 0.72 $\pm$ 0.16 & 0.73 $\pm$ 0.21 \\
Run 4 & 0.73 $\pm$ 0.18 & 0.72 $\pm$ 0.17 \\
Run 5 & 0.74 $\pm$ 0.21 & 0.73 $\pm$ 0.19 \\
\bottomrule
\end{tabular}
}}
\end{table}
Table~\ref{tab:stability} demonstrates that our approach yields stable and reproducible results. While minor variations in generated code may occur, the extracted features maintain consistent discriminative power.

\subsection{Qualitative Analysis (RQ5)}

\begin{figure}[t]
\centering
\includegraphics[width=0.9\linewidth]{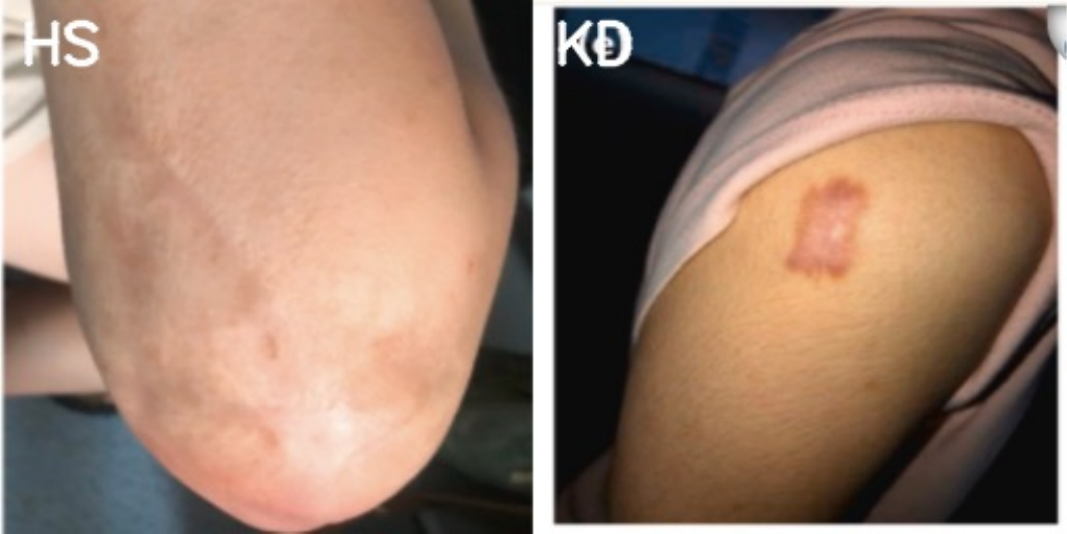}
\caption{Visualization of clinically meaningful features. (a) Hypertrophic scar samples with confined morphology and (b) Keloid samples with high erythema and irregular boundaries. Feature values align with established clinical criteria.}
\label{fig:qualitative}
\end{figure}

To illustrate interpretability, Figure~\ref{fig:qualitative} visualizes representative scar images alongside their extracted features. Since our data comes from hospital collections, patient privacy is a factor that prevents visualization. To clarify this, we searched for publicly available papers on PubMed and collected images specifically for interpretability verification.
The feature values align with clinical expectations: keloid samples exhibit higher erythema scores (increased redness) and lower solidity values (irregular boundaries extending beyond the wound), while hypertrophic scars show more confined morphology.
This transparency enables clinicians to validate model predictions against their domain expertise, facilitating trust and adoption in clinical workflows.

\section{Conclusion}

We presented a novel framework that repositions large language models as knowledge-driven feature engineers for medical image classification. Rather than training data-hungry neural networks end-to-end or using LLMs as black-box classifiers, our approach leverages LLMs to generate executable feature extraction code grounded in established clinical criteria. This design explicitly decouples domain knowledge acquisition from statistical learning, enabling effective classification in data-scarce medical domains.

Extensive experiments on pathological scar classification demonstrate that our method consistently outperforms both end-to-end deep learning models and direct multimodal LLM classification under limited data conditions. The extracted features are inherently interpretable, aligning with clinical assessment scales such as the Vancouver Scar Scale, which facilitates expert validation and clinical trust. Furthermore, by processing images through locally executed code, our approach preserves patient privacy without transmitting sensitive data to external servers.

\paragraph{Limitations and Future Work.}
Our current framework relies on a single LLM generation step; future work could explore iterative refinement based on classification feedback or ensemble multiple generated feature extractors. The approach also assumes access to a capable LLM, which may introduce dependency on proprietary models. Investigating open-source alternatives or fine-tuned domain-specific LLMs would enhance accessibility. Additionally, while we focused on scar classification, the framework is general and could be extended to other medical imaging tasks such as skin lesion analysis, histopathology, and radiology, which we leave for future investigation.

\paragraph{Broader Impact.}
This work contributes to the growing effort of developing trustworthy AI for healthcare. By providing interpretable, privacy-preserving, and data-efficient alternatives to black-box deep learning, we hope to lower barriers for AI adoption in clinical practice, particularly in specialized domains where expert data is scarce. We encourage the research community to further explore the paradigm of using LLMs as knowledge externalizers rather than direct decision-makers in high-stakes applications.

\paragraph{Usage of LLMs.}
Large Language Models were used as assistive tools in the preparation of this manuscript. We employed LLMs for grammar checking, LaTeX formatting, and improving the clarity of technical descriptions.  
The core scientific contributions and conclusions presented in this paper originate from the authors' work.

\paragraph{Ethics Statement}
All patient images were collected with informed consent under hospital-approved protocols and used solely for research purposes. Data handling followed clinical privacy guidelines.

\bibliographystyle{named}
\bibliography{ijcai26_refined}

\end{document}